\documentclass{article}
\usepackage{amssymb}

\usepackage{graphicx}
\usepackage{xcolor}
\usepackage{amsmath}
\usepackage{multicol}
\usepackage{multirow}
\usepackage{subcaption}
\usepackage{wrapfig}

\usepackage{tabularx}
\usepackage{booktabs}
\usepackage{pifont}
\usepackage{caption}
\usepackage{floatflt}
\usepackage[bookmarks=true]{hyperref}
\usepackage[usenames,dvipsnames,table,xcdraw]{xcolor}
\definecolor{tablepeach}{RGB}{255, 240, 235}

\newcommand{\cmark}{\textcolor{blue}{\ding{51}}}%
\newcommand{\xmark}{\textcolor{red}{\ding{55}}}%

\usepackage[final]{corl_2025} 

\newcommand{\acronym}{HAT}
\newcommand{\datasetname}{PH$^2$D}

\title{Humanoid Policy $\sim$ Human Policy}

\author{
  Ri-Zhao Qiu\textsuperscript{\rm *,1} \quad 
  Shiqi Yang\textsuperscript{\rm *,1} \quad
  Xuxin Cheng\textsuperscript{\rm *,1} \quad
  Chaitanya Chawla\textsuperscript{\rm *,2} \quad
  Jialong Li\textsuperscript{\rm 1} \quad \\
  \textbf{Tairan He}\textsuperscript{\rm 2} \quad
  \textbf{Ge Yan}\textsuperscript{\rm 4} \quad
  \textbf{David Yoon}\textsuperscript{\rm 3} \quad
  \textbf{Ryan Hoque}\textsuperscript{\rm 3} \quad 
  \textbf{Lars Paulsen}\textsuperscript{\rm 1} \quad \\
  \textbf{Ge Yang}\textsuperscript{\rm 5} \quad
  \textbf{Jian Zhang}\textsuperscript{\rm 3} \quad
  \textbf{Sha Yi}\textsuperscript{\rm 1} \quad
  \textbf{Guanya Shi}\textsuperscript{\rm 2} \quad
  \textbf{Xiaolong Wang}\textsuperscript{\rm 1} \quad \\
  \textsuperscript{\rm 1} UC San Diego,  
  \textsuperscript{\rm 2} CMU,
  \textsuperscript{\rm 3} Apple,
  \textsuperscript{\rm 4} University of Washington,
  \textsuperscript{\rm 5} MIT \\
  \url{https://human-as-robot.github.io/}
}

\begin{document}
\maketitle

\begin{center}
    \centering
    \captionsetup{type=figure}
    \vspace{-8pt}
    \includegraphics[width=.94\textwidth]{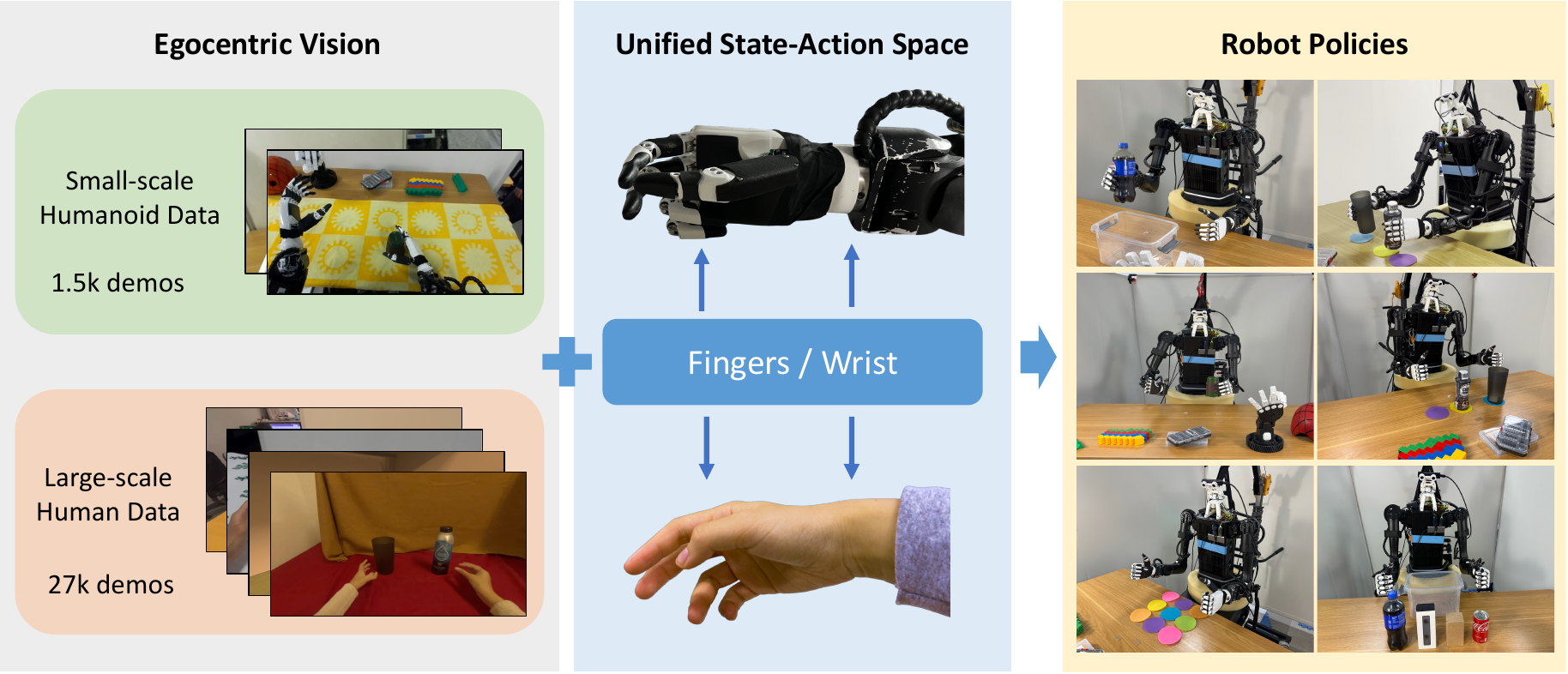}
    \vspace{-4pt}
    \caption{This paper advocates high-quality human data as a data source for cross-embodiment learning - \textbf{task-oriented} egocentric human data. We collect a large-scale dataset, \textbf{P}hysical \textbf{H}uman-\textbf{H}umanoid \textbf{D}ata (\textbf{\datasetname{}}), with hand-finger 3D poses from consumer-grade VR devices on well-defined manipulation tasks directly aligned with robots. Without relying on modular perception, we train a Human Action Transformer ({\bf HAT}) manipulation policy by directly modeling humans as a different humanoid embodiment in an end-to-end manner.}
    \label{fig:teaser}
\end{center}


\begin{abstract}
    Training manipulation policies for humanoid robots with diverse data enhances their robustness and generalization across tasks and platforms. However, learning solely from robot demonstrations is labor-intensive, requiring expensive tele-operated data collection, which is difficult to scale. This paper investigates a more scalable data source, egocentric human demonstrations, to serve as cross-embodiment training data for robot learning. We mitigate the embodiment gap between humanoids and humans from both the data and modeling perspectives. We collect an egocentric \textbf{task-oriented} dataset (\textbf{\datasetname{}}) that is directly aligned with humanoid manipulation demonstrations. We then train a human-humanoid behavior policy, which we term Human Action Transformer (\textbf{\acronym{}}). The state-action space of \acronym{} is unified for both humans and humanoid robots and can be differentiably retargeted to robot actions. Co-trained with smaller-scale robot data, \acronym{} directly models humanoid robots and humans as different embodiments without additional supervision. We show that human data improve both generalization and robustness of \acronym{} with significantly better data collection efficiency.
\end{abstract}



\section{Introduction}
\label{sec:intro}

Learning from real robot demonstrations has led to great progress in robotic manipulation recently~\cite{liu2024-rdt,ghosh2024-octo,black2024-pi0,dasari2024-DP}. One key advancement to enable such progress was hardware / software co-designs to scale up data collection using teleoperation~\cite{zhao2023learning,fu2024-mobilealoha,chi2024universal,yang2024ace,cheng2024-opentv,he2024-omnih2o} and directly controlling the robot end effector~\cite{dasari2019robonet,bharadhwaj2024roboagent,zhao2023learning,fu2024-mobilealoha,ha2024-umi-on-legs,chi2024universal}. Instead of gathering data on a single robot, collective efforts have been made to merge diverse robot data and train foundational policies across embodiments~\cite{dasari2019robonet,2023-openxembodiment,ghosh2024-octo,liu2024-rdt,black2024-pi0,dasari2024-DP}, which have shown to improve cross-embodiment and cross-task generalizability.
\begin{floatingtable}[r]{
  \begin{tabular}{l cccc}
    \toprule
    \multirow{2}{*}{Dataset} & \multicolumn{2}{c}{Human} & \multicolumn{2}{c}{Robot}\\
    \cmidrule(lr){2-3} \cmidrule(lr){4-5}
    & \# Frames & \# Demos & \# Frames & \# Demos \\
    \midrule
    DexCap~\cite{wang2024-dexcap} & $\sim$378k & 787 &  NA & NA \\
    EgoMimic~\cite{kareer2024-egomimic} & $\sim$432k$^\dagger$ & 2,150 & \textbf{1.29M}$^\dagger$ & 1,000 \\
    \datasetname{} (Ours) & $\sim$\textbf{3.02M} & \textbf{26,824} & $\sim$668k & \textbf{1,552} \\
    \bottomrule
    \end{tabular}
}
\vspace{-5pt}
\caption{\textbf{Comparisons of task-oriented egocentric human datasets.} Besides having the most demonstrations, \datasetname{} is collected on various manipulation tasks, diverse objects and scenes, with accurate 3D hand-finger poses and language annotations. $^\dagger$: estimated based on reported data collection time with 30\,Hz; whereas DexCap~\cite{wang2024-dexcap} and \datasetname{} report processed frames for training.}
\label{tab:dataset_comp}
\end{floatingtable}

However, collecting structured real-robot data is expensive and time-consuming. We are still far away from building a robust and generalizable model as what has been achieved in Computer Vision~\cite{radford2021-CLIP} and NLP~\cite{openai2023gpt4}. If we examine humanoid robot teleoperation more closely, it involves robots mimicking human actions using geometric transforms or retargeting to control robot joints and end-effectors. From this perspective, \textbf{we propose to model robots in a human-centric representation}, and the robot action is just a transformation away from the human action. If we can accurately capture the end-effector and head poses of humans, egocentric human demonstrations will be a more scalable source of training data, as we can collect them efficiently, in any place, and without a robot.

In this paper, we perform cross-human and humanoid embodiment training for robotic manipulation. Our key insight is to model bimanual humanoid behaviors by \textit{directly imitating human behaviors without using learning surrogates} such as affordances~\cite{mendonca2023-human-world,bahl2023-humanaffordances}. To realize this, we first collect an egocentric task-oriented dataset of \textbf{P}hysical \textbf{H}umanoid-\textbf{H}uman \textbf{D}ata, dubbed \datasetname{}. We adapt consumer-grade VR devices to collect egocentric videos with automatic but accurate hand pose and end effector ({\it i.e.,} hand) annotations. Compared to existing human daily behavior datasets~\cite{grauman2022-ego4d,Damen2018-epickitchen}, \datasetname{} is task-oriented so that it can be directly used for co-training. The same VR hardwares are then used to perform teleoperation to collect smaller-scale humanoid data for better alignment. We then train a Human-humanoid Action Transformer (HAT), which predicts future hand-finger trajectories in a unified human-centric state-action representation space. To obtain robot actions, we simply apply inverse kinematics and hand retargeting to differentiably convert human actions to robot actions for deployment.

We conduct real-robot evaluations on different manipulation tasks with extensive ablation studies to investigate how to best align human and humanoid demonstrations. In particular, we found that co-training with diverse human data improves robustness against spatial variance and background perturbation, generalizing in settings unseen in robot data but seen in human data. We believe that these findings highlight the potential of using human data for large-scale cross-embodiment learning.

In summary, our contributions are:
\begin{itemize}
    \item \textbf{A dataset}, \datasetname{}, which is a large egocentric, task-oriented human-humanoid dataset with accurate hand and wrist poses for modeling human behavior (see Tab.~\ref{tab:dataset_comp}).
    \item \textbf{A cross human-humanoid manipulation policy}, \acronym{}, that introduces a unified state-action space and other alignment techniques for humanoid manipulation.
    \item \textbf{Improved policy robustness and generalization} validated by extensive experiments and ablation studies to show the benefits of co-training with human data.
\end{itemize}

\section{Related Work}
\label{sec:related_work}

\textbf{Imitation Learning for Robot Manipulation.} Recently, learning robot policy with data gathered directly from the multiple and target robot embodiment has shown impressive robustness and dexterity~\cite{zhao2024-aloha-unleashed,ghosh2024-octo,wang2024-HPT,liu2024-rdt, chi2023diffusion, qiu2024-wildlma, cheng2024-opentv, lu2024-mobiletv,ze2024-idp3}. The scale of data for imitation learning has grown substantially with recent advancements in data collection~\cite{arunachalam2023dexterous, cheng2024-opentv, chi2024universal, yang2024ace}, where human operators can efficiently collect large amounts of high-quality, task-oriented data. Despite these advances, achieving open-world generalization still remains a significant challenge due to lack of internet-scale training data.

\textbf{Learning from Human Videos.} Learning policies from human videos is a long-standing topic in both computer vision and robotics due to the vast existence of human data. Existing works can be approximately divided into two categories: aligning observations or actions.

\textbf{Learn from Human - Aligning Observations.} While teleoperating the actual robot platform allows learning policy with great dexterity, there is still a long way to go to achieve higher levels of generalization across diverse tasks, environments, and platforms. Unlike fields such as computer vision~\cite{radford2021-CLIP} and natural language processing~\cite{openai2023gpt4} benefiting from  internet-scale data, robot data collection in the real world is far more constrained. Various approaches have attempted to use internet-scale human videos to train robot policies~\cite{chen2021learning, lee2017learning, lee2013syntactic, nguyen2018translating, rothfuss2018deep, yang2015robot}. Due to various discrepancies ({\it e.g.,} supervision and viewpoints) between egocentric robot views and internet videos, most existing work~\cite{mendonca2023-human-world,bahl2023-humanaffordances} use modular approaches with intermediate representations as surrogates for training. The most representative ones are affordances~\cite{mendonca2023-human-world,bahl2023-humanaffordances} for object interaction, object keypoints predictions~\cite{bharadhwaj2024-track2act,wen2023-anypointtraj,li2024-okami, das2021model, xiong2021learning}, or other types of object representations~\cite{pirk2019online, nair2022r3m, ma2022vip}. 

\textbf{Learn from Human - Aligning Actions.} Beyond observation alignment, transferring human demonstrations to robotic platforms introduces additional challenges due to differences in embodiment, actuation, and control dynamics. Specific alignment of human and robot actions is required to overcome these disparities. Approaches have employed masking in egocentric views~\cite{kareer2024-egomimic}, aligning motion trajectories or flow~\cite{lin2024flowretrieval, ren2025motion}, object-centric actions~\cite{zhu2024vision, hsu2024spot}, or hand tracking with specialized hardware~\cite{wang2024-dexcap}. Most closely related to our work, HumanPlus~\cite{fu2024-humanplus} designs a remapping method from 3D human pose estimation to tele-operate humanoid robots. Compared to HumanPlus, the insight of our method is to waive the requirement for robot hardware in collecting human data and collect diverse human data directly for co-training. In contrast to HumanPlus, we intentionally avoid performing retargeting on human demonstrations and designed the policy to directly use human hand poses as states/actions. On the other hand, the `human shadowing' retargeting in HumanPlus is a teleoperation method that still requires robots, leading to lower collection efficiency than ours.

\textbf{Cross-Embodiment.} Cross-embodiment pre-training has been shown to improve adaptability and generalization over different embodiments~\cite{huang2020one, chen2024mirage, yang2024pushing, yang2023polybot, ebert2021bridge, franzmeyer2022learn, ghadirzadeh2021bayesian, shankar2022translating, xu2023xskill, yin2022cross, zakka2022xirl, zhang2021policy, zhang2020learning}. When utilizing human videos, introducing intermediate representations can be prone to composite errors. Recent works investigate end-to-end approaches~\cite{ghosh2024-octo,wang2024-HPT,liu2024-rdt,black2024-pi0} using cross-embodied robot data to reduce such compounding perceptive errors. Noticeably, these works have found that such end-to-end learning leads to desired behaviors such as retrying~\cite{black2024-pi0}. Some other work~\cite{bahl2022-humanimitation,li2024-okami} enforces viewpoint constraints between training human demonstrations and test-time robot deployment to allow learning on human data but it trades off the scalability of the data collection process.

\textbf{Concurrent Work.} 
Some concurrent work~\cite{wang2024-dexcap,kareer2024-egomimic,wang2023-mimicplay} also attempts to use egocentric human demonstrations for end-to-end cross-embodiment policy learning. DexCap~\cite{wang2024-dexcap} uses gloves to track 3D hand poses with a chest-mounted RGBD camera to capture egocentric human videos. However, DexCap relies on 3D inputs, whereas some recent works~\cite{black2024-pi0,liu2024-rdt} have shown the scalability of 2D visual inputs. Most related to our work, EgoMimic~\cite{kareer2024-egomimic} also proposes to collect data using wearable device~\cite{engel2023-project-aria} with 2D visual inputs. However, EgoMimic requires strict visual sensor alignments; whereas we show that scaling up diverse observations with different cameras makes the policy more robust. In addition, \datasetname{} is also greater in dataset scale and object diversity. We also show our policy can be deployed on real robots without strict requirements of visual sensors and heuristics, which paves the way for scalable data collection.

\section{Method}

To collect more data to train generalizable robot policies, recent research has explored cross-embodiment learning, enabling policies to generalize across diverse physical forms~\cite{black2024-pi0,liu2024-rdt,dasari2024-DP,ghosh2024-octo,khazatsky2024-droid,2023-openxembodiment}. This paper proposes egocentric human manipulation demonstrations as a scalable source of cross-embodiment training data. Sec.~\ref{sec:dataset} describes our approach to adapt consumer-grade VR devices to scale up human data collection conveniently for a dataset of task-oriented egocentric human demonstrations. Sec.~\ref{sec:hat} describes various techniques to handle domain gaps to align human data and robot data for learning humanoid manipulation policy.

\subsection{\datasetname{}: Task-oriented \textbf{P}hysical \textbf{H}umanoid-\textbf{H}uman \textbf{D}ata}
\label{sec:dataset}

Though there has been existing work that collects egocentric human videos~\cite{kareer2024-egomimic,Damen2018-epickitchen,grauman2022-ego4d,wang2024-dexcap}, they either (1) provide demonstrations mostly for non-task-oriented skills ({\it e.g., dancing}) and do not provide world-frame 3D head and hand poses estimations for imitation learning supervision~\cite{grauman2022-ego4d,Damen2018-epickitchen} or (2) require specialized hardware or robot setups~\cite{wang2024-dexcap,kareer2024-egomimic}.

\begin{wrapfigure}{r}{0.5\textwidth}
    \centering
    \vspace{-16pt}
    \includegraphics[width=\linewidth]{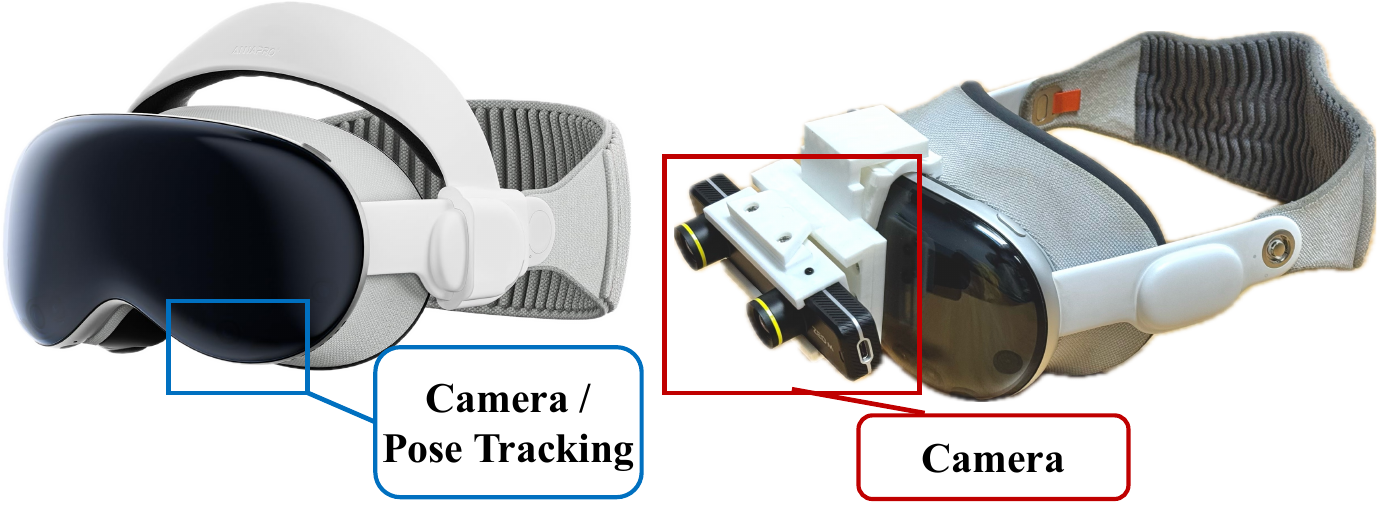}
    \caption{\small{\textbf{Consumer-grade Devices for Data Collection.} To avoid relying on specialized hardware for data collection to make our method scalable, we design our data collection process using consumer-grade VR devices.}}
    \label{fig:hardware_setup}
    \vspace{-16pt}
\end{wrapfigure}

To address these issues, we propose \datasetname{}. \datasetname{} address these two issues by (1) collecting task-oriented human demonstrations that are directly related to robot execution, (2) adapting well-engineered SDKs of VR devices (illustrated in Fig.~\ref{fig:hardware_setup}) to provide supervision, and (3) diversifying tasks, camera sensors, and reducing whole-body movement to reduce domain gaps in both vision and behaviors.

\paragraph{Adapting Low-cost Commerical Devices} With development in pose estimation~\cite{zhu2023-motionbert} and system engineering, modern mobile devices are capable of providing accurate on-device world frame 3D head pose tracking and 3D hand keypoint tracking~\cite{cheng2024-opentv}, which has proved to be stable enough to teleoperate robot in real-time~\cite{cheng2024-opentv,ha2024-umi-on-legs}. We design software and hardware to support convenient data collection across different devices. Different cameras provide better visual diversity.
\begin{itemize}
  \item \textbf{Apple Vision Pro + Built-in Camera.} We developed a Vision OS App that uses the built-in camera for visual observation and uses the Apple ARKit for 3D head and hand poses.
  \item \textbf{Meta Quest 3 / Apple Vision Pro + ZED Camera.} We developed a web-based application based on OpenTelevision~\cite{cheng2024-opentv} to gather 3D head and hand poses. We also designed a 3D-printed holder to mount ZED Mini Stereo cameras on these devices. This configuration is both low-cost ($<$700\$) and introduces more diversity with stereo cameras.
\end{itemize}

\paragraph{Data Collection Pipeline}
We collect task-oriented egocentric human demonstrations by asking human operators to perform tasks overlapping with robot execution ({\it e.g.,} grasping and pouring) when wearing the VR devices. For every demonstration, we provide language instructions (\textit{e.g., grasp a can of coke zero with right hand}), and synchronize proprioception inputs and visual inputs by closest timestamps.

\textbf{Action Domain Gap.} Human actions and tele-operated robot actions exhibit two distinct characteristics: (1) human manipulation usually involves involuntary whole-body movement, and (2) humans are more dexterous than robots and have significantly faster task completion time than robots. We mitigate the first gap by requesting the human data collectors to sit in an upright position. For the second speed gap, we interpolate translation and rotations of human data during training (effectively `slowing down' actions). The slow-down factors $\alpha_{\text{slow}}$ are obtained by normalizing the average task completion time of humans and humanoids, which is empirically distributed around $4$. For consistency, we use $\alpha_{\text{slow}} = 4$ in all tasks.

\begin{figure*}
  \centering
  \includegraphics[width=0.98\linewidth]{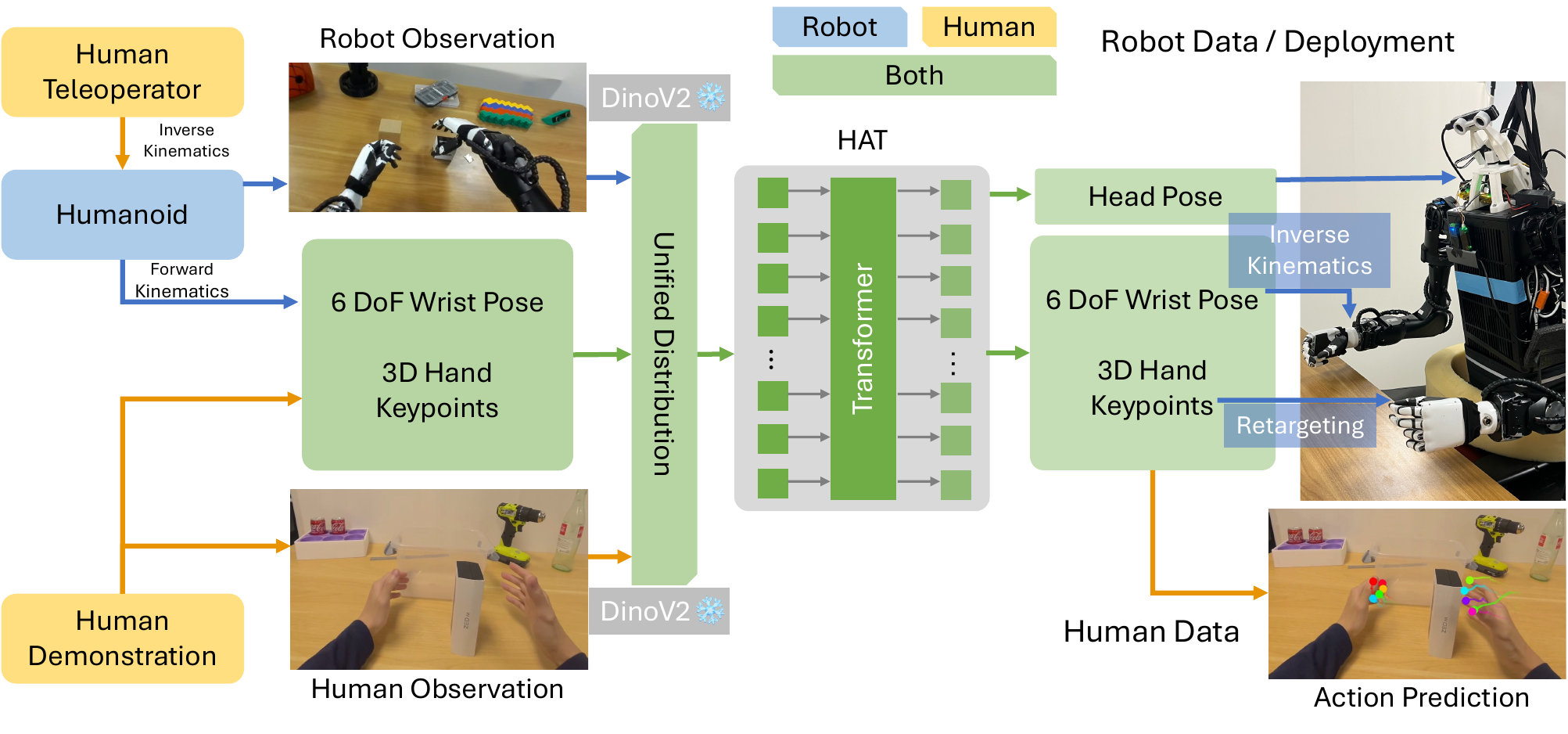}
  \caption{\textbf{Overview of \acronym{}.} Human Action Transformer (\acronym{}) learns a robot policy by modeling humans. During training, we sample a state-action pair from either human data or robot data. The images are encoded by a frozen DinoV2 encoder~\cite{oquab2023-dinov2}. The \acronym{} model makes predictions in a human-centric action space using wrist 6 DoF poses and finger tips, which is retargeted to robot poses during real-robot deployment.}
  \label{fig:hat_overview}
  \vspace{-10pt}
\end{figure*}

\subsection{\acronym{}: Human Action Transformer}
\label{sec:hat}

\acronym{} learns cross-embodied robot policy by modeling humans. We demonstrate that treating bimanual humanoid robots and humans as different robot embodiments via retargeting improves both generalizability and robustness of \acronym{}.

More concretely, let $\mathcal{D}_{robot} = \{(\mathbf{S}_i, \mathbf{A}_i)\}_{i = 1}^{N}$ be the set of data collected from real bimanual humanoid robots using teleoperation~\cite{cheng2024-opentv}, where $\mathbf{S}_i$ is the states including proprioceptive and visual observations of $i$-th demonstration and $\mathbf{A}_i$ be the actions. The collected \datasetname{} dataset, $\mathcal{D}_{human} = \{(\mathbf{\tilde{S}}_i, \mathbf{\tilde{A}}_i)\}_{i = 1}^{M}$ is used to augment the training process. Note that it is reasonable to assume $M \gg N$ due to the significantly better human data collection efficiency.

The goal is to design a policy $\pi: \mathbf{S} \to \mathbf{A}$ that predicts future robot actions $\mathbf{a}_t$ given current robot observation $\mathbf{s}_t$ at time $t$, where the future actions $\mathbf{a}_{t + 1}$ is usually a chunk of actions for multi-step execution (with slight abuse of notation). We model $\pi$ using \acronym{}, which is a transformer-based architecture predicting action chunks~\cite{zhao2023learning}. The overview of the model is illustrated in Fig.~\ref{fig:hat_overview}. We discuss key design choices of \acronym{} with experimental ablations.

\textbf{Unified State-Action Space.} Both bimanual robots and humans have two end effectors. In our case, our robots are also equipped with an actuated 2DoF neck that can rotate, which resembles the autonomous head movement when humans perform manipulation. Therefore, we design a unified state-action space ({\it i.e.,} $(\mathbf{S}, \mathbf{A}) \equiv (\mathbf{\tilde{S}}, \mathbf{\tilde{A}})$) for both bimanual robots and humans. More concretely, the proprioceptive observation is a 54-dimensional vector (6D rotations~\cite{zhou2019-6drot} of the head, left wrist, and right wrist; x/y/z of left and right wrists and 10 finger tips). In this work, since we deploy our policy on robots with 5-fingered dexterous hands (shown in Fig.~\ref{fig:humanoid_hardware}), there exists a bijective mapping between the finger tips of robot hands and human hands. Note that injective mapping is also possible ({\it e.g.,} mapping distance between the thumb finger and other fingers to parallel gripper distance).

\textbf{Visual Domain Gap.} Two types of domain gaps exist for co-training on human/humanoid data: camera sensors and end effector appearance. Since our human data collection process includes cameras different from robot deployment, this leads to camera domain gaps such as tones. Also, the appearances of human and humanoid end effectors are different. However, with sufficiently large and diverse data, we find it not a strict necessity to apply heuristic processing such as visual artifacts~\cite{kareer2024-egomimic} or generative methods~\cite{yu2023-scaling} to train human-robot policies - basic image augmentations such as color jittering and Gaussian blurring are effective regularization.

\textbf{Training.} The final policy is denoted as $\pi: f_{\theta}(\cdot) \to \mathbf{A}$ for both human and robot policy, where $f_{\theta}$ is a transformer-based neural network parametrized by $\theta$. The final loss is given by,
\begin{equation}
    \mathcal{L} = \ell_1(\pi(s_i), a_i) + \lambda \cdot \ell_1(\pi(s_i)_{\text{EEF}}, a_{i, \text{EEF}})\,,
\end{equation}
where $\text{EEF}$ are the indices of the translation vectors of the left and right wrists, and $\lambda = 2$ is an (insensitive) hyperparameter used to balance loss to emphasize the importance of end effector positions over learning unnecessarily precise finger tip keypoints.

\section{Experiments}
\label{sec:experiments}

\begin{table*}[t]
\centering
\resizebox{\linewidth}{!}{ 
  \begin{tabular}{c c c c c c c c c c c c c}
\toprule
\multirow{2}{*}{Meth.} & \multirow{2}{*}{H. Data} 
& \multirow{2}{*}{D. Norm} &
\multicolumn{2}{c}{Passing} & \multicolumn{2}{c}{Horizontal Grasp} & \multicolumn{2}{c}{Vertical Grasp} & \multicolumn{2}{c}{Pouring} & \multicolumn{2}{c}{Ovr. Succ.} \\
\cmidrule(lr){4-5} \cmidrule(lr){6-7} \cmidrule(lr){8-9} \cmidrule(lr){10-11} \cmidrule(lr){12-13}
 & & & I.D. & O.O.D. & I.D. & O.O.D. & I.D. & O.O.D.  & I.D. & O.O.D. & I.D. & O.O.D. \\
\midrule
 ACT
    &  \xmark &  NA 
    & 19/20 & 36/60 
    & 8/10 & 7/30
    & 7/20 & 15/70
    & \textbf{8/10} & 1/10
    & 42/60 & 59/170
 \\
 HAT
    & \cmark & \xmark 
    & 17/20  & 51/60
    & \textbf{9/10} & 11/30
    & \textbf{14/20} & \textbf{30/70}
    & 5/10 & 5/10
    & 45/60 & 97/170
 \\
 HAT
    & \cmark & \cmark 
    & \textbf{20/20} & \textbf{52/60}  
    & 8/10 & \textbf{12/30}
    & 13/20 & 29/70
    & \textbf{8/10} & \textbf{8/10}
    & \textbf{49/60} & \textbf{101/170}
 \\
 \midrule
 \multicolumn{3}{c}{\textbf{Type of Generalization}} & \multicolumn{2}{c}{Background} & \multicolumn{2}{c}{Texture} & \multicolumn{2}{c}{Obj. Placement}
 \\
 \bottomrule
 \end{tabular}
 }
 \caption{\textbf{Success rate of autonomous skill execution}. Co-training with human data (H. Data) significantly improves the Out-Of-Distribution (O.O.D.) performance with nearly 100\% relative improvement on all tasks on Humanoid A. We also ablate the design choice of using different normalizations (D. Norm) for different embodiments. We designate each task setting to investigate a single type of generalization. Detailed analysis of each type of generalization is presented in Sec.~\ref{sec:individual_generalization_analysis}.
 } 
 \label{table:main_results}
 \vspace{-5pt}
\end{table*}

\begin{wrapfigure}{r}{0.5\textwidth}
    \centering
    \vspace{-20pt}
    \includegraphics[width=\linewidth]{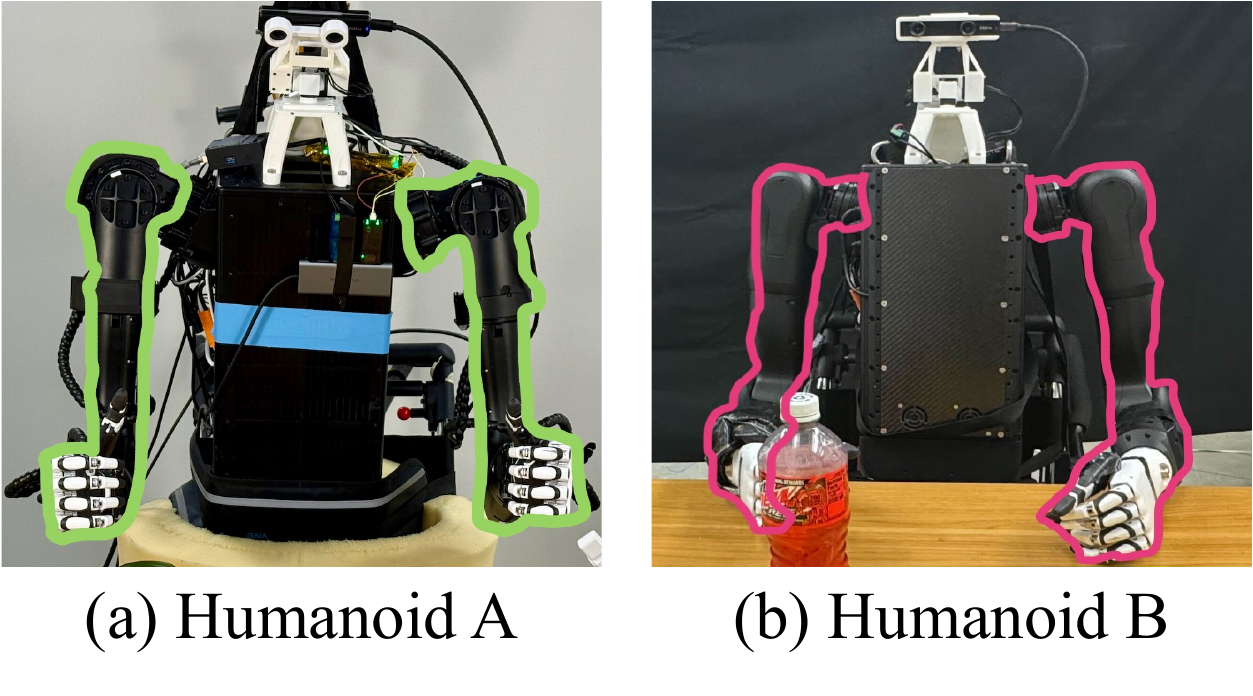}
    \vspace{-15pt}
    \caption{\small{\textbf{Hardware Illustration.} Most robot data attributes to Humanoid A, a Unitree H1 robot. Humanoid B, a Unitree H1-2 robot with \textbf{different} arm motor configurations, is used to evaluate few-shot cross-humanoid transfer. Detailed comparisons in Sec.~\ref{sec:humanoid_differences}}}
    \label{fig:humanoid_hardware}
    \vspace{-20pt}
\end{wrapfigure}

\paragraph{Hardware Platforms.} We run our experiments on two humanoid robots (Humanoid A and Humanoid B shown in Fig.~\ref{fig:humanoid_hardware}) equipped with 6-DOF Inspire dexterous hands. Humanoid A is a Unitree H1 robot and Humanoid B is a Unitree H1\_2 robot with different arm configurations. Similar to humans, both robots (1) are equipped with actuated necks~\cite{cheng2024-opentv} to get make use of egocentric views and (2) do not have wrist cameras. Unless otherwise noted, most humanoid data collection is done with Humanoid A. We use Humanoid B mainly for testing cross-humanoid generalization.

\paragraph{Implementation Details.} We implement policy architecture by adopting an transformer-based architecture predicting future action chunks~\cite{zhao2023learning}. We use a frozen DinoV2 ViT-S~\cite{oquab2023-dinov2} as the visual backbone. We implement two variants: (1) \textbf{ACT}: baseline implementation using the Action Chunk Transformer~\cite{zhao2023learning}, trained using only robot data. Robot states are represented as joint positions. (2) \textbf{\acronym{}}: same architecture as ACT, but the state encoder operates in the unified state-action space. Unless otherwise stated, HAT is co-trained on robot and human data. A checkpoint is trained for each task with approximately 250-400 robot demonstrations.

\paragraph{Experimental Protocol.} We collect robot and human demonstrations in different object sets. Since human demonstrations are easier to collect, the settings in human demonstrations are generally more diverse, which include background, object types, object positions, and the relative position of the human to the table.

\begin{figure}[t!]%
\begin{subfigure}[t]{0.49\linewidth}
\includegraphics[width=\textwidth]{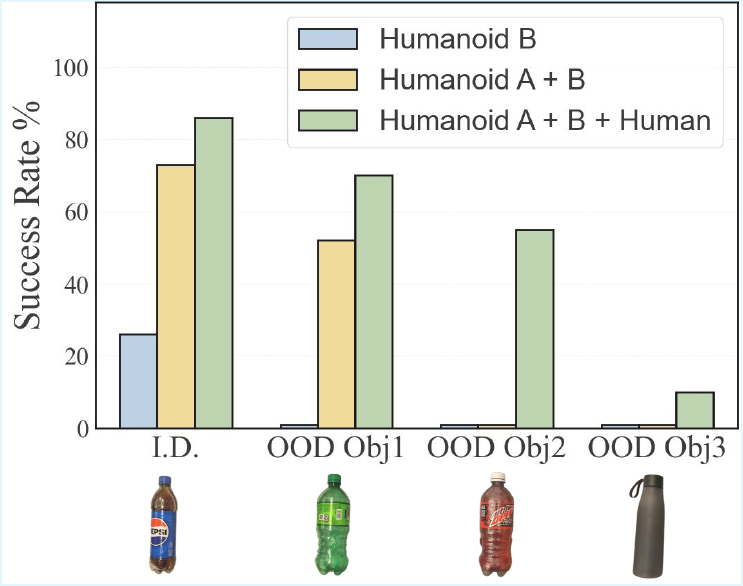}
\caption{Performance of Humanoid B co-trained with \datasetname{} on horizontal grasping. o1 is seen by Humanoid B. o2 and o3 seen in human data. o4 is unseen in all data.}
\label{fig:object_data_cmu}
\end{subfigure}
\hfill%
\begin{subfigure}[t]{0.49\linewidth}
\includegraphics[width=\textwidth]{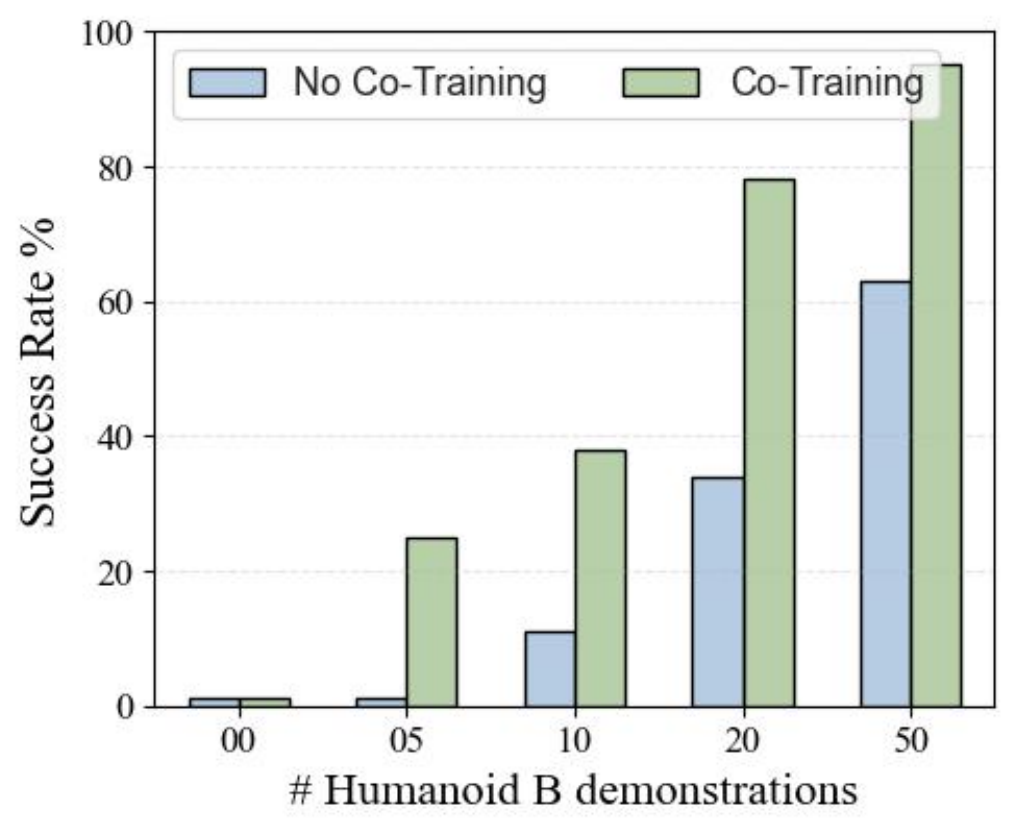}
\caption{Co-training consistently outperforms isolated training as Humanoid B demonstrations increase, achieving good success rates even in low-data regimes.}
\label{fig:ours_rotation}
\end{subfigure}
\caption{\textbf{Few-Shot Adaptation.}
    Co-training consistently outperforms isolated training as Humanoid B demonstrations increase, achieving robust success rates even in low-data regimes.}
    \label{fig:co-training_cmu}
\vspace{-16pt}
\end{figure}

We experimented with four different dexterous manipulation tasks and investigated in-distribution and out-of-distribution setups. The \textit{in-distribution (I.D.)} setting tests the learned skills with backgrounds and object arrangements approximately similar to the training demonstrations presented in the real-robot data. In the Out-Of-Distribution (O.O.D.) setting, we test generalizability and robustness by introducing novel setups that were presented in human data but not in robot data. Fig.~\ref{fig:task_illustration} visualizes different manipulation tasks and how we define out-of-distribution settings for each task.

\subsection{Main Evaluation}

\noindent{\textbf{Human data has minor effects on I.D. testing.}} From Tab.~\ref{table:main_results}, we can see that I.D. performance with or without co-training with human data gives similar results. In the I.D. setting, we closely match the scene setups as training demonstrations, including both background, object types, and object placements. Thus, policies trained with only a small amount of Humanoid A data performed well in this setting. This finding is consistent with recent work~\cite{cheng2024-opentv,chi2024universal} that frozen visual foundation models~\cite{radford2021-CLIP,oquab2023-dinov2} improve robustness against certain external perturbations such as lighting.

\noindent{\textbf{Human data improves the O.O.D. settings with many generalizations.}} One common challenge in imitation learning is overfitting to only in-distribution task settings. 
Hence, it is crucial for a robot policy to generalize beyond the scene setups seen in a limited set of single-embodiment data.
To demonstrate how co-training with human data reduces such overfitting, we introduce O.O.D. task settings to evaluate such generalization. From Tab.~\ref{table:main_results}, we can see that co-training drastically improves O.O.D. settings, achieving nearly 100\% relative improvement in settings unseen by the robot data. In particular, we find that human data improves three types of generalization: \textbf{background, object placement, and appearance}. To isolate the effect of each variable, each task focuses on a specific type of generalization as listed in Tab.~\ref{table:main_results}, with in-depth analyses in Sec.~\ref{sec:individual_generalization_analysis}.

\subsection{Few-Shot Transfer across Heterogenous Embodiments}

We conducted few-shot generalization experiments on a distinct humanoid platform (Humanoid B), contrasting it with our primary platform, Humanoid A. Notably, Humanoid B’s demonstration data were collected in an entirely separate environment, introducing both embodiment and environmental shifts. We highlight two key advantages of our approach: (1) the ability to unify heterogeneous human-centric data sources (humanoids and humans) into a generalizable policy framework, and (2) the capacity to rapidly adapt to new embodiments with drastically reduced data requirements.

\begin{table}[t]
\begin{minipage}{.4\textwidth}
    \centering
    \includegraphics[width=\linewidth]{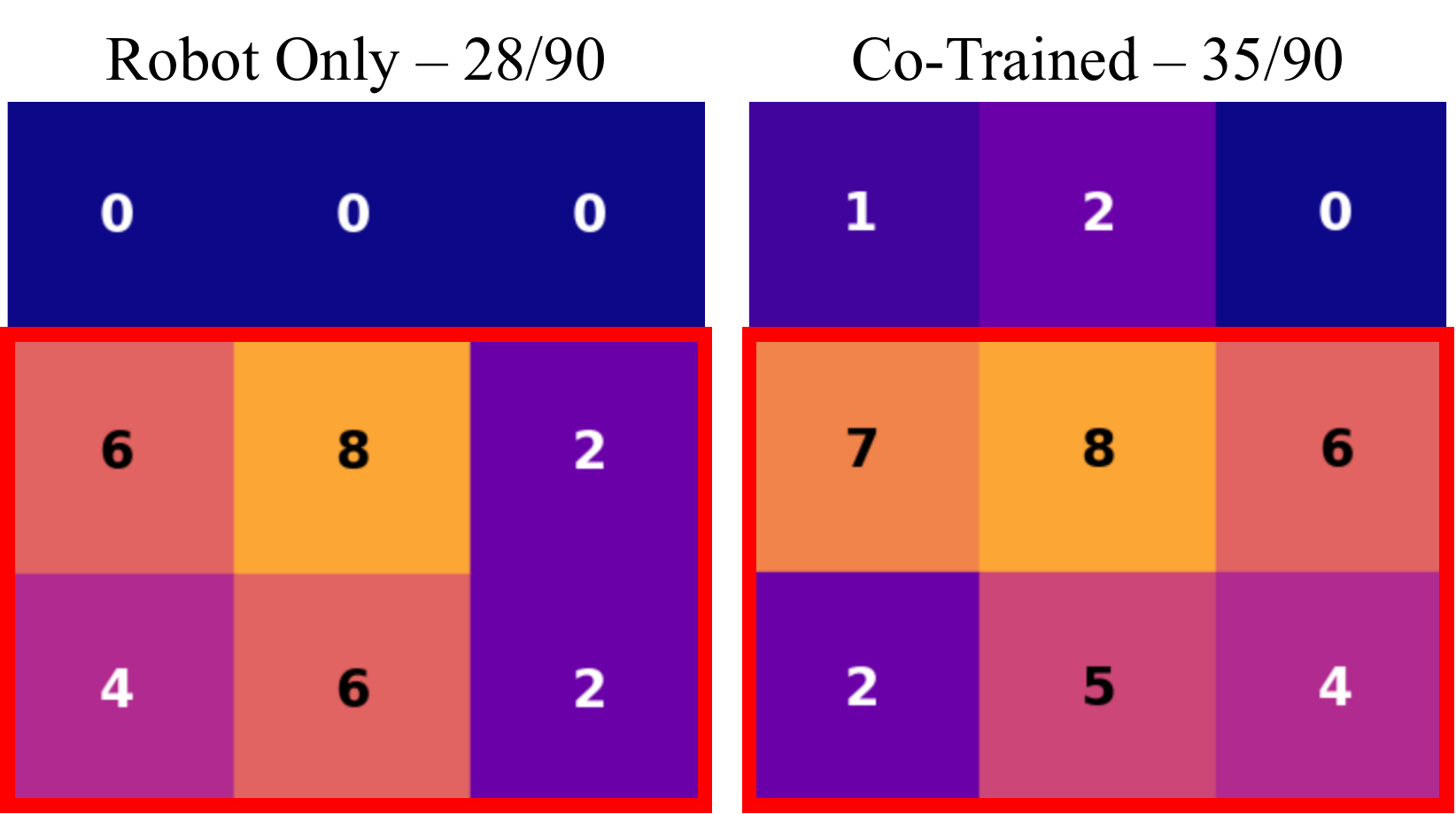}
    \captionof{figure}{\textbf{Human data has better sampling efficiency}. Per-grid vertical grasping successes out of 10 trials with models trained with robot-only data and mixed data. Red boxes indicate where training data is collected.}
    \label{fig:sampling_efficiency}
\end{minipage}
\hspace{8px}
\begin{minipage}{.59\textwidth}
  \label{tab:rendering_tab}
  \centering
  \resizebox{\linewidth}{!}{
      \begin{tabular}{c c c c}
     \toprule
     Task & State Space & Action Speed & Success \\
    \midrule
     & \cmark  & \xmark &  1/10 \\
     Vertical Grasping & \xmark  & \cmark &  0/10 \\
     & \cmark  & \cmark &  \textbf{4/10} \\
    \bottomrule
    \end{tabular}
  }
  \caption{\textbf{Importance of unifying policy inputs and outputs.} We report the number of successes of vertical grasping objects in the upper-left block as illustrated in Fig.~\ref{fig:human_position}. Baselines use joint positions as state input or do not interpolate human motions.}
  \label{table:state_speed_ablation}
\end{minipage}
\vspace{-16pt}
\end{table}

\textit{Experiment 1: Cross-embodiment co-training efficacy}
Using only 20 demonstrations from Humanoid B, we trained 3 policies - respectively on data from (i) Humanoid B only, (ii) Humanoid B + Humanoid A (cross-embodiment), and (iii) Humanoid B + Humanoid A + Human (cross-embodiment and human priors). As shown in Fig.~\ref{fig:object_data_cmu}, co-trained policies (ii) and (iii) substantially outperformed the Humanoid B-only baselines on all task settings, underscoring the method’s ability to transfer latent task structure across embodiments.

\textit{Experiment 2: Scaling Demonstrations for Few-Shot Adaptation} 
We further quantified the relationship between required for few-shot generalization. We hold Humanoid A and human datasets fixed for the horizontal grasping task and ablate number of demonstrations required for Humanoid B in Fig.~\ref{fig:co-training_cmu}. Co-training (Humanoid B + A + Human) consistently outperformed isolated training on Humanoid B across all settings, especially in the few-data regime.

\subsection{Ablation Study}

\noindent{\textbf{Sampling Efficiency of Human and Humanoid Data.}}  Conceptually, collecting human data is less expensive, not just because it can be done faster, but also because it can be done in in-the-wild scenes; reduces setup cost before every data collection; and avoids the hardware cost to equip every operator with robots.

We perform additional experiments to show that even in the lab setting, human data can have better sampling efficiency in unit time. In particular, we provide a small-scale experiment on the vertical grasping task. Allocating 20 minutes for two settings, we collected (1) 60 Humanoid A demonstrations, (2) 30 Humanoid A demonstrations, and 120 human demonstrations. To avoid conflating diversity and data size, the object placements in all demonstrations are \underline{evenly distributed} at the bottom 6 cells. The results are given in Fig.~\ref{fig:sampling_efficiency}. The policy trained with mixed robot and human data performs significantly better, which validates the sampling efficiency of human data over robot data. Each cell represents a 10cm × 10cm region where the robot attempts to pick up a box.

\noindent{\textbf{State-Action Design.}} In Tab.~\ref{table:state_speed_ablation}, we ablate the design choices of the proprioception state space and the speed of output actions. In particular, using the same set of robot and human data, we implement two baselines: 1) a unified state-action space, but does not interpolate ({\it i.e.,} slow down) the human actions; and 2) a baseline that interpolates human actions but uses separate state representation for humanoid (joint positions) and humans (EEF representation). The policies exhibit different failure patterns during the rollout of these two baselines. Without interpolating human actions, the speed of the predicted actions fluctuates between fast (resembling humans) and slow (resembling teleoperation), which leads to instability. 
Without a unified state space, the policy is given a `shortcut' to distinguish between embodiments, which leads to on-par in-distribution performance and significantly worse OOD performance.

\noindent{\textbf{More Ablation Study.}} Due to space limit, please refer to the appendix and the supplementary for more qualitative visualization and quantitative ablation studies.

\section{Conclusions}
\label{sec:conclusion}

This paper proposes \datasetname{}, an effort to construct a large-scale human task-oriented behavior dataset, along with the training pipeline \acronym{}, which leverages \datasetname{} and robot data to show how humans can be treated as a data source for cross-embodiment learning. We show that it is possible to directly train an imitation learning model with mixed human-humanoid data without any training surrogates when the human data are aligned with the robot data. The learned policy shows improved generalization and robustness compared to the counterpart trained using only real-robot data.


\section{Limitations}
Although we also collect language instructions in \datasetname{}, due to our focus on investigating the embodiment gap between humans and humanoids, one limitation of the current version of the paper uses a relatively simple architecture for learning policy. In the near future, we plan to expand the policy learning process to train a large language-conditioned cross-embodiment policy to investigate generalization to novel language using human demonstrations. The collection of human data relies on off-the-shelf VR hardwares and their hand tracking SDKs. Since these SDKs were trained mostly for VR applications, hand keypoint tracking can fail for certain motions with heavy occlusion. In addition, though the proposed method conceptually extends to more robot morphologies, current evaluations are done on robots equipped with dexterous hands.

\section{Acknowledgment}

This work was supported, in part, by NSF CAREER Award IIS-2240014, NSF CCF-2112665 (TILOS), and gifts from Amazon, Meta and Apple.


\clearpage


\bibliography{main}  

\clearpage

\appendix

\begin{figure*}[!t]
    \centering
    \begin{subfigure}[t]{\textwidth}
    \centering
    \includegraphics[width=\textwidth]{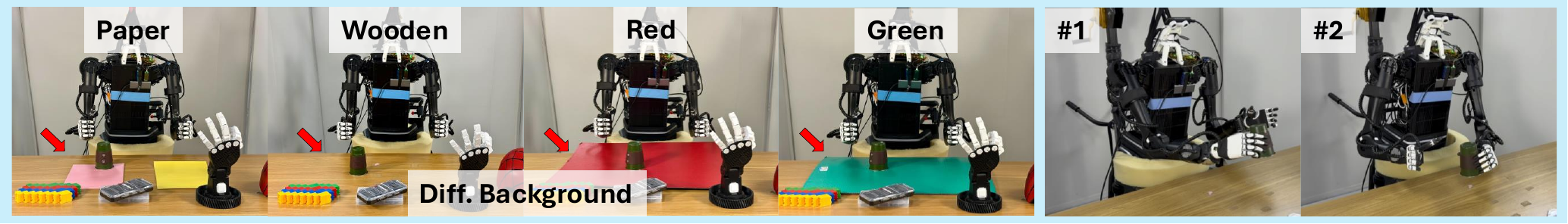}
    \caption{The robot performs the \textbf{cup passing }task across four different backgrounds. The left side shows the four background variations, while the right side illustrates the two passing directions: (\textit{\#1} - Right hand passes the cup to the left hand, \textit{\#2} - Left hand passes the cup to the right hand).}
    \vspace{5pt}
    
    \label{fig:passing}
    \end{subfigure} 
    
    \begin{subfigure}[t]{\textwidth}
    \centering
    \includegraphics[width=\textwidth]{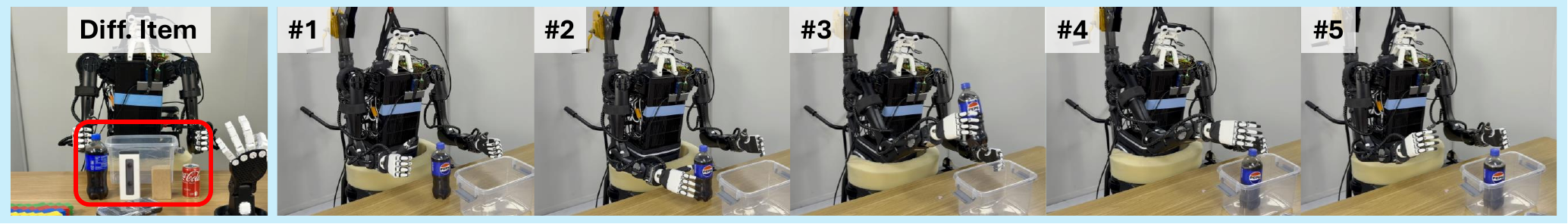}
    \caption{The robot performs the \textbf{horizontal grasping} task with four different items: bottle, box\_1, box\_2, and can, as shown on the left. The right side illustrates the process: (\textit{\#1-\#3} - The robot grasps the bottle, \textit{\#4-\#5} - The robot places it into the plastic bin).}
    \vspace{5pt}
    
    \label{fig:grasping}
    \end{subfigure} 

    \begin{subfigure}[t]{\textwidth}
    \centering
    \includegraphics[width=\textwidth]{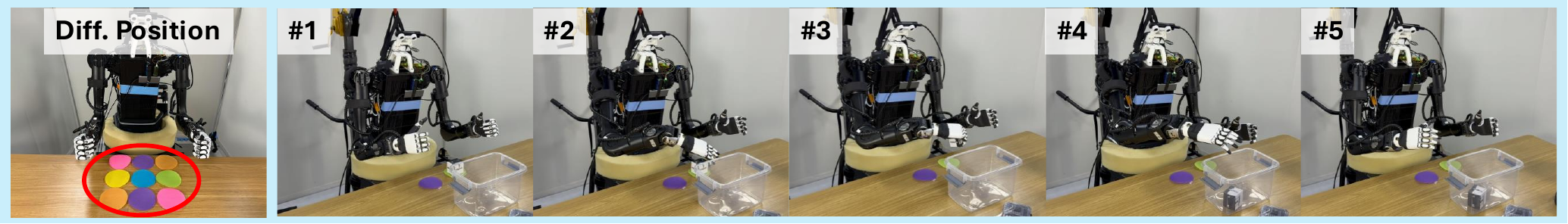}
    \caption{The robot performs the \textbf{vertical grasping} task. As shown on the left, the Dynamixel box is placed in nine different positions for grasping. The right side illustrates the process: (\textit{\#1-\#3} - The robot grasps the box, \textit{\#4-\#5} - The robot places the box into the plastic bin).}
    \vspace{5pt}
    
    \label{fig:picking}
    \end{subfigure}
    
    \begin{subfigure}[t]{\textwidth}
    \centering
    \includegraphics[width=\textwidth]{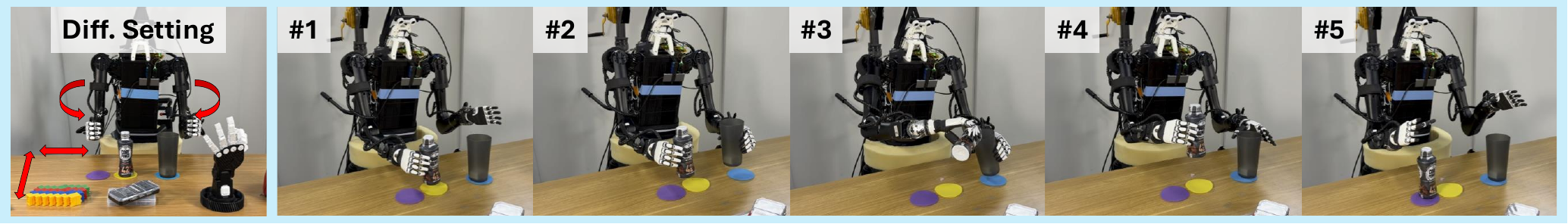}
    \caption{The robot performs the \textbf{pouring} task. The left side shows different settings achieved by varying the robot's rotation and the table's position. The right side illustrates the pouring process: (\textit{\#1} - Right hand grasps the bottle, \textit{\#2} - Left hand grasps the cup, \textit{\#3} - Pouring the drink, \textit{\#4} - Left hand places the cup down, \textit{\#5} - Right hand places the bottle down).}

    \label{fig:pouring}
    \end{subfigure}
    
   \caption{Illustrations of tasks used in quantitative evaluations. From top to bottom: cup passing, horizontal grasping, vertical grasping, and pouring.}
    \label{fig:task_illustration}
    \vspace{-5pt}
\end{figure*}

\section{More Ablation Study - Data Collection}

\begin{table}[t]
  \centering
 \begin{tabular}{c c c c c c}
 \toprule
 \multirow{2}{*}{Method}  & Bottle & Box$_{1}$ & Box$_2$ & Can
  & \multirow{2}{*}{Ovr. Succ.} \\
\cmidrule(lr){2-5} 
   & I.D. & H.D. & H.D. & H.D. &  \\
\midrule
Without whole-body  & 8/10 &  \textbf{6/10} & 0/10 & \textbf{7/10} & \textbf{21/40} \\
With whole-body  & 9/10 &  3/10 & \textbf{3/10} & 3/10 & 18/40 \\
\bottomrule
\end{tabular}
\caption{\textbf{Ablation of how human whole-body movement in training demonstrations affects policy rollout}. We collect the same number of demonstrations on the same set of objects for the {\it grasping} task with or without whole-body movement. Since the robot does not have a natural whole-body movement like humans, it negatively influences the manipulation success rate.}
\vspace{-10pt}
\label{table:whole_body}
\end{table}

\noindent{\textbf{Autonomous Whole-body Movement.}} In Tab.~\ref{table:whole_body}, we justify the necessity to minimize body movement in human data collection. Humans tend to move their upper body unconsciously during manipulation (including shoulder and waist movement). However, existing humanoid robots have yet to reach such a level of dexterity. Thus, having these difficult-to-replicate actions in the human demonstrations leads to degraded performance. We hypothesize that such a necessity would be greatly reduced with the development of both whole-body locomotion methods and mechanical designs, but for the currently available platforms, we instruct operators to minimize body movement as much as possible in our dataset.

\noindent{\textbf{Efficiency of Data Collection.}} 
In Tab.~\ref{table:demo_efficiency_comp}, we compare task completion times across different setups, including standard human manipulation, human demonstrations performed while wearing a VR device, and robot teleoperation. This analysis highlights how task-oriented human demonstrations can be a scalable data source for cross-embodiment learning. Notably, wearing a VR device does not significantly impact human manipulation speed, as the completion time remains nearly the same as in standard human demonstrations.

Among different data collection schemes, we find that most overhead arises during the retargeting process from human actions to robot actions. This is primarily due to latency and the constrained workspace of 7-DoF robotic arms, which are inherent challenges in existing data collection methods such as VR teleoperation~\cite{cheng2024-opentv}, motion tracking~\cite{fu2024-humanplus,he2024-omnih2o}, and puppeting~\cite{yang2024ace,zhao2023learning}.

Beyond data collection speed, human demonstrations offer several additional advantages over teleoperation. They provide a safer alternative, reducing risks associated with real-robot execution. They are also more labor-efficient, as they do not require additional personnel for supervision. Furthermore, human demonstrations allow for greater flexibility in settings, enabling a diverse range of environments without requiring robot-specific adaptations. Additionally, human demonstrations achieve a higher demonstration success rate, and the required hardware (such as motion capture or VR devices) is more accessible and cost-effective compared to full robotic setups. These factors collectively make human data a more scalable solution for large-scale data collection.

\begin{table}[t]
\centering
\begin{tabular}{l c c}
\toprule
Method & Grasping (secs) & Pouring (secs) \\
\midrule
Human Demo                   & 3.79$\pm$0.27 & 4.81$\pm$0.35 \\
\textbf{Human Demo with VR}   & 4.09$\pm$0.30 & 4.90$\pm$0.26 \\
Humanoid Demo (VR Teleop)     & 19.72$\pm$1.65 & 37.31$\pm$6.25 \\
\bottomrule
\end{tabular}
\caption{\textbf{Amortized mean and standard deviation of the time required to collect a single demonstration}, including scene resets. The first row shows the time for regular human to complete corresponding tasks in real world. The second row represents our human data when wearing VR for data collection, demonstrating that egocentric human demonstrations provide a more scalable data source compared to robot teleoperation.}
\label{table:demo_efficiency_comp}
\vspace{-5pt}
\end{table}

\section{Normalization of different embodiments.} 
\begin{figure}[t]
  \centering
  \includegraphics[width=0.6\linewidth]{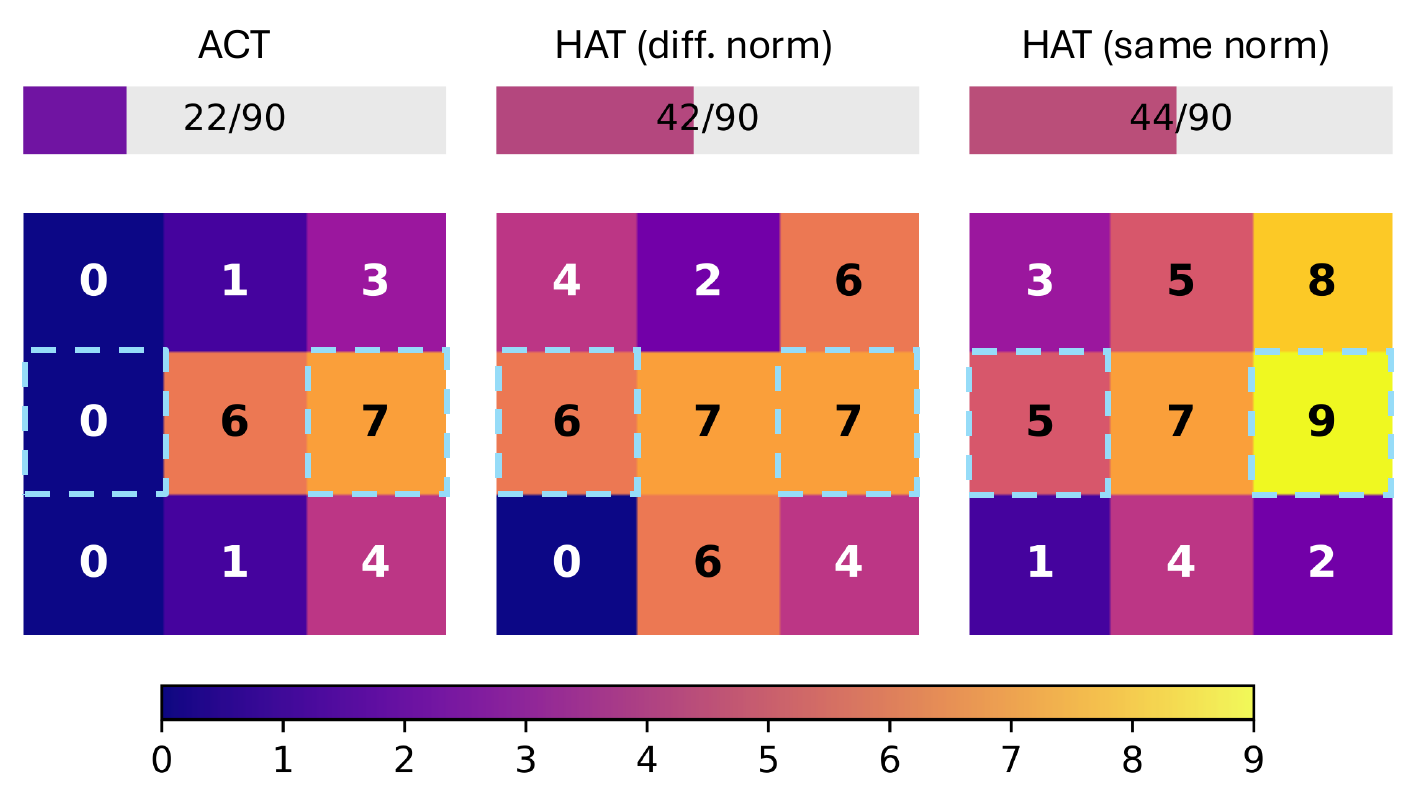}
  \caption{\textbf{Object Placement Generalization.}
    Performance comparisons of models trained with and without human data on vertical grasping (picking). Each cell in the 3×3 grid represents a 10cm × 10cm region where the robot attempts to pick up a box, with numbers indicating successful attempts out of 10. The real-robot data is collected in two cells inside the dashed lines. Notably, our teleoperation data is intentionally imbalanced. 
    }
  \label{fig:human_position}
\end{figure}

Tab.~\ref{table:main_results} suggests minor differences between using different normalization coefficients for the states and actions vectors of humans and humanoids. We take a closer look in Fig.~\ref{fig:human_position}, where we investigate the impact of different normalization strategies in the vertical grasping (picking) task. Noticeably, the same normalization approach achieved the highest overall success rate, but the success distribution is biased towards the upper-right region of the grid.

We hypothesize that this is because humans have a larger workspace than humanoid robots. Thus, human data encompasses humanoid proprioception as a subset, which results in a relatively smaller distribution for the robot state-action space.

\section{In-Depth Analysis of Different Types of Generalization}
\label{sec:individual_generalization_analysis}

\noindent{\textbf{Human data improves background generalization.}} We chose to use the \textit{cup passing} task to test background generalization. We prepared four different tablecloths as backgrounds, as shown in Fig.~\ref{fig:passing}. In terms of training data distribution, the teleoperation data for this task was collected exclusively on the paper background shown in Fig.~\ref{fig:passing}, whereas the human data includes more than five different backgrounds. This diverse human dataset significantly enhances the generalization ability of the co-trained HAT policy. As shown in Tab.~\ref{table:background_general}. , HAT consistently outperforms across all four backgrounds, demonstrating robustness to background variations. In addition, the overall success rate increases by nearly 50\% compared to training without human data, highlighting the advantage of utilizing diverse human demonstrations.

\noindent{\textbf{Human data improves appearance generalization.}} To test how co-training improves robustness to perturbations in object textures, we evaluate the \textit{horizontal grasping} policy on novel objects, as shown in Fig.~\ref{fig:grasping}. Specifically, we compare the policy’s performance on the bottle, box\_1, box\_2, and can, as shown left to right in the first image in Fig.~\ref{fig:grasping}. These objects differ significantly in both color and shape from the bottle used in the teleoperation data distribution.

Since grasping is a relatively simple task, our adjusted policy demonstrates strong learning capabilities even with only 50 teleoperation data samples. The policy can successfully grasp most bottles despite the limited training set. To better highlight the impact of human data, we selected more challenging objects for evaluation. As shown in Tab.~\ref{table:item_general}, human data significantly enhances the policy’s ability to grasp these more difficult objects.

Notably, box\_1 appears in the human data, while box\_2 does not. Despite this, we observe that co-training with human data still improves overall performance, even on box\_2, though its success rate does not increase. This suggests that, beyond direct experience with specific objects, the human data helps the policy learn broader visual priors that enable more proactive and stable grasping behaviors. For box\_2, while the success rate remains low—partially due to its low height and color similarity to the table—the co-trained HAT policy demonstrates fewer out-of-distribution (OOD) failures and more actively searches for graspable regions. The failures on box\_2 are primarily due to unstable grasping and the small box slipping from the hand, rather than the inability to perceive or locate the object.

Furthermore, adding more human data not only improves performance on objects seen in human training demonstrations (e.g., box\_1) but also enhances generalization to completely novel objects (e.g., box\_2 and can). We hypothesize that, as the number of objects grows, \acronym{} starts to learn inter-category visual priors that guide it to grasp objects more effectively, even when they were not explicitly present in the training set.

\noindent{\textbf{Human data improves object placement generalization.}}
Finally, we introduce variations in object placements that are not present in the real-robot training demonstrations and specifically investigate this in the \textit{vertical grasping (picking)} task. In this task, we intentionally constrain the robot data collection to object placements within a subset of cells, while human vertical grasping data covers a much more diverse range of settings.

To systematically analyze the impact of human data, we evaluate model performance on a structured 3×3 grid, where each cell represents a 10cm × 10cm region for grasping attempts. The numbers in each cell indicate the number of successful picks out of 10 trials. Real-robot training data is collected from only two specific cells, highlighted with dashed lines.

A key detail in our teleoperation data distribution is that 50 picking attempts are collected from the right-hand side grid and only 10 from the left-hand side grid. This imbalance explains why policies trained purely on teleoperation data struggle to grasp objects in the left-side grid. We observe that models trained solely on robot data fail to generalize to unseen cells, whereas cross-embodiment learning with human data significantly improves generalization, doubling the overall success rate.

\begin{table}[t] 
  \centering
 \begin{tabular}{c c c c c c}
 \toprule
 \multirow{2}{*}{Method}  & Bottle & Box$_{1}$ & Box$_2$ & Can
  & \multirow{2}{*}{Ovr. Succ.} \\
\cmidrule(lr){2-5} 
   & I.D. & H.D. & O.O.D. & O.O.D. &  \\
\midrule
ACT & 8/10 & 5/10 & \textbf{1/10} & 1/10 & 16/40 \\
HAT & 8/10 & \textbf{7/10} & \textbf{1/10} & \textbf{4/10} & \textbf{21/40} \\
\bottomrule
\end{tabular}

\caption{\textbf{Object Appearance Generalization:} In the horizontal grasping task, we evaluated the grasping performance by attempting to grasp each object 10 times and recorded the success rate. 
}
\label{table:item_general}
\vspace{-10pt}
\end{table}

\begin{table}[t]
\centering 
 \begin{tabular}{c c c c c c}
\toprule
\multirow{2}{*}{Method}  & Paper & Wooden & Red & Green
  & \multirow{2}{*}{Ovr. Succ.} \\
\cmidrule(lr){2-5} 
   & I.D. & H.D. & O.O.D. & O.O.D. &  \\
\midrule
ACT & 19/20 & 14/20 & 12/20 & 10/20 & 55/80 \\
HAT  & \textbf{20/20} & \textbf{16/20} & \textbf{18/20} & \textbf{18/20} & \textbf{72/80} \\
\bottomrule
\end{tabular}
\caption{\textbf{Background Generalization:} In the cup passing task, we evaluate the passing performance by recording the number of failures or retries needed to complete 20 cup-passing trials. 
}
\label{table:background_general}
\end{table}

\section{In-Depth Comparison between Humanoid A and Humanoid B configurations}
\label{sec:humanoid_differences}

This section presents a detailed comparison of the two humanoid platforms, referred to as Humanoid A and Humanoid B, with a focus on joint structure and implications for manipulation capabilities. We restrict our analysis to the arm configurations, as other parts of the body were not exclusively explored in this work.

While morphologically similar, these two humanoids have drastically different arm configurations that create hurdles in direct policy transfer. Besides differences in motor technical specs such as torque and types of encoder (Humanoid B has absolute motor position encoders), they also have different mechanical limits. The range of motion (ROM) for the first four proximal joints—shoulder\_pitch, shoulder\_roll, shoulder\_yaw, and elbow—differs across the two platforms. Humanoid B exhibits a consistently wider ROM, which allows a wider set of reachable configurations and increases the manipulability of the arm in constrained environments. Table \ref{table:rom_comparison} summarizes the ROM values for these shared joints.

A more significant architectural divergence is observed at the wrist. Humanoid A includes a single distal joint—wrist\_roll—providing limited wrist articulation. This restricts end-effector dexterity and constrains in-hand manipulation strategies to a single rotational degree of freedom. In contrast, Humanoid B is equipped with a complete wrist mechanism composed of three independently actuated joints: wrist\_pitch, wrist\_roll, and wrist\_yaw. These additional degrees of freedom allow for full orientation control of the end-effector, enabling tasks that require precise alignment, rotation, and fine adjustment of object poses.

\begin{table}[h]
\centering
\begin{tabular}{|l|c|c|}
\hline
\textbf{Joint} & \textbf{Humanoid A} & \textbf{Humanoid B} \\
\hline
shoulder\_pitch & $-164^\circ$ to $+164^\circ$ & $-180^\circ$ to $+90^\circ$   \\
shoulder\_roll  & $-19^\circ$ to $+178^\circ$ & $-21^\circ$ to $+194^\circ$     \\
shoulder\_yaw   & $-74^\circ$ to $+255^\circ$ & $-152^\circ$ to $+172^\circ$    \\
elbow           & $-71^\circ$ to $150^\circ$ & $-54^\circ$ to $182^\circ$     \\
wrist\_roll           & $-175^\circ$ to $175^\circ$ & $-172^\circ$ to $157^\circ$     \\
\hline
\end{tabular}
\caption{Joint Range of Motion Comparison between Humanoid A and B (in degrees)}
\label{table:rom_comparison}
\vspace{-10pt}
\end{table}

\end{document}